# Evaluating Personal Information Output from Conversational Interactions in Generative AI Systems

Yosuke Seki
*Faculty of Foreign Studies*
*Kobe City University of Foreign Studies*
Kobe, Japan
seki@inst.kobe-cufs.ac.jp

Hirotaka Tahara
*Department of Systems and Information Engineering*
*Kobe City College of Technology*
Kobe, Japan
h-tahara@kobe-kosen.ac.jp

***Abstract*—This exploratory pilot study evaluates the scope and perceived accuracy of personal information output from ongoing conversational interactions in generative AI systems using GPT-5.2 Instant and GPT-5.2 Thinking, categorized into three output types—Fact, Inference, and Confidence. Based on the evaluation results obtained from 15 Japanese participants, differences in model design have limited impact on personal information output tendencies. Compared with the Inference type, the Fact type shows a more conservative output pattern. Regarding attribute categories, the findings indicate that Core Personal attributes associated with identification are treated relatively conservatively, whereas Behavioral and Linguistic attributes show higher accuracy across both Fact and Inference outputs. Furthermore, Holistic Profile, Psychological and Cognitive, and Residual attributes are more readily inferred, even when not supported by explicit factual outputs. Notably, the lack of null outputs for these attributes in the Inference type suggests that such inferred profiles may be constructed from indirectly available contextual information. The findings may contribute to future discussions regarding privacy awareness and personal information inference in generative AI systems.**

***Keywords—generative AI, privacy, personal information***

## I. Introduction

Generative AI systems, such as ChatGPT, Gemini, and DeepSeek, which are based on large language models (LLMs), have become widely used in various domains, including education, healthcare, business, and everyday communication [1]. These systems are typically designed to generate responses based on user-provided prompts. Generative AI systems that incorporate long-term memory can generate responses based on past interaction histories. As a result, there are growing concerns about the potential for generative AI to output personal information from seemingly innocuous prompt inputs [2].

Recent studies have shown that LLMs are capable of inferring a wide range of personal attributes from textual inputs alone. Staab et al. demonstrated that modern LLMs can accurately infer demographic and socioeconomic attributes, such as age, gender, location, and occupation, from collections of user-authored documents, even when explicit identifiers are removed [2]. Their findings indicate that subtle linguistic cues enable large-scale user profiling, highlighting privacy risks beyond simple memorization.

Similarly, Yukhymenko et al. systematically evaluated personal attribute inference using collections of social media or forum-style posts generated by synthetic users [3]. By controlling user attributes within a synthetic dataset, their study showed that inference accuracy varies depending on the type of personal attribute, suggesting that some attributes are more easily inferred than others.

In addition to demographic and socioeconomic attributes, psychological characteristics have also been shown to be inferable from textual inputs. Peters and Matz reported that LLMs can predict users' personality traits based on social media posts with accuracy comparable to that of specialized supervised models [4]. Their results raise ethical concerns, as such attributes may be inferred without users' awareness.

However, despite these advances, existing studies primarily focus on inference from collections of user-authored documents, such as social media posting histories, or on specific categories of personal attributes. Consequently, the extent to which generative AI systems can output different types of personal information in interactive conversational settings has not yet been sufficiently clarified. Such investigations have become feasible due to recent advances in long-term memory and LLMs. To address this gap, this exploratory pilot study evaluates the scope and perceived accuracy of personal information that generative AI systems can output from ongoing conversational interactions, without relying on explicitly provided external personal data.

The contributions of this study are as follows.

- Personal information outputs are classified into Fact, Inference, and Confidence using GPT-5.2 Instant and GPT-5.2 Thinking, and the scope, accuracy, and null tendencies are evaluated using a prompt procedure.
- Personal information is organized into multiple attribute categories, and differences in output tendencies across attributes are clarified.
- The relationship between Inference and Confidence is analyzed to examine the extent to which generative AI expresses confidence in inferred personal attributes.

## II. Study Design

### *A. Selection of Generative AI System*

According to a survey conducted by LINE Research [5], which targeted male and female respondents aged 15 to 69 across Japan, ChatGPT accounted for 74.6% of generative AI usage as of June 2025. Although usage rates may have changed at the time of writing, ChatGPT is likely the most widely used

Author preprint of a paper presented at IIAI-AAI 2026. Published version: DOI 10.23919/IIAI-AAICPS00095.2026.00071.

generative AI system in Japan; therefore, it was selected as the target system in this study.

ChatGPT is available in both free and paid versions. Although the specific mechanisms and duration of memory retention are not fully disclosed, the paid version provides enhanced memory capabilities that enable the system to retain and utilize information derived from past interactions across sessions. In contrast, such functionality is more limited in the free version. Therefore, this study focuses on the paid version of ChatGPT.

As of January 2026, the paid version of ChatGPT allows users to select between two model variants: GPT-5.2 Instant and GPT-5.2 Thinking. GPT-5.2 Instant is designed to prioritize response speed and is characterized by its ability to generate answers within a relatively short time. In contrast, GPT-5.2 Thinking is intended to perform more multi-step internal reasoning and tends to produce more careful and detailed responses. Because it is not known in advance how these model differences may affect the content and accuracy of personal information inference, this study employs both models for investigation.

### B. *Prompt Design for Personal Information Output*

To obtain users' personal information, it is necessary to carefully examine the content of the prompts. Personal information that can be obtained includes information that may be explicitly stated during interactions with a conversational AI system. For example, during casual conversations, users may mention their gender or place of origin, and it is likely that such information is recorded by generative AI systems. In addition, there are personal attributes that can be implicitly inferred from the dialogue content. For instance, when a user frequently handles academic information using a generative AI system, it may be possible to infer that the user's occupation is that of a university faculty member or a researcher. Therefore, personal information was categorized into three output types—Fact, Inference (attribute-level outputs), and Confidence (a 1-6 score)—and prompts were designed to elicit each type separately. Requiring the generative AI system to assign a Confidence level to each Inference attribute enables quantitative analysis of both the inferred personal information and the reliability of those inferences.

The set of personal information was designed to cover a wide range of attributes commonly discussed in studies on user profiling and personal information inference [6][7]. In this study, the attributes were organized into higher-level conceptual attribute categories—Core Personal (Items 1–7), Behavioral (Items 8–9), Linguistic (Items 10–11), Professional (Items 12–14), Psychological and Cognitive (Items 15–16), and Residual (Item 17). In addition, a Holistic Profile (Item 18) was included to capture integrative impressions that transcend individual attributes. Residual includes outputs that did not clearly fit into the other predefined categories. Holistic Profile refers to outputs describing comprehensive impressions or overall characterizations of participants that cannot be sufficiently represented by individual attribute categories alone. The prompt used in the investigation is as follows:

> [Content] Based on the conversation so far, please summarize my inferred profile for each of the attributes listed below. Clearly distinguish between facts (i.e., information that I have explicitly stated) and inferences (i.e., information that you have inferred). If an attribute cannot be determined from the conversation, respond with "Not applicable."
>
> [List of attributes] 1. Name; 2. Age; 3. Gender; 4. Place of origin; 5. Family composition; 6. Economic status; 7. Current residence; 8. Interests (hobbies, favorite topics, frequently discussed domains); 9. Lifestyle and behavioral traits; 10. Languages used; 11. Language proficiency; 12. Occupation/Position; 13. Technical literacy; 14. Field of expertise (topics handled or studied at work or school); 15. Personality; 16. Values and worldview; 17. Other; 18. Overall inferred profile.
>
> [Output format] For each attribute, describe a set consisting of Fact / Inference / Confidence. Indicate the confidence level of each inference on a scale from 1 (low) to 6 (high).

## III. RESULTS

The participants consisted of 15 Japanese individuals, recruited through a crowdsourcing platform, who had used the paid version of ChatGPT for more than six months. Their mean age was 36.60 years (SD = 11.54). The mean weekly usage frequency was 6.4 times per week (SD = 1.25; range: 0–7). Each participant was asked to input the designed prompt and then evaluate the model's Fact and Inference outputs based on their subjective judgements using a six-point Likert scale ranging from 1 (not correct) to 6 (correct), with intermediate values indicating gradations of perceived accuracy. It is important to note that Confidence scores were assigned by the model itself, whereas Fact and Inference scores were evaluated by human participants; therefore, the interpretation of these results requires caution.

### A. *Comparison of the Scores*

Figure 1 shows the proportions of score distributions for each model. Proportions of approximately 5% or less are omitted. The proportions of high scores (5 or 6) were 77.88% for GPT-5.2 Instant and 76.02% for GPT-5.2 Thinking in the Fact type, 71.34% and 72%, respectively, in the Inference type, and 60.25% and 60.44% in the Confidence type. These results indicate that there were no substantial differences between the two models in terms of the proportions of high scores.

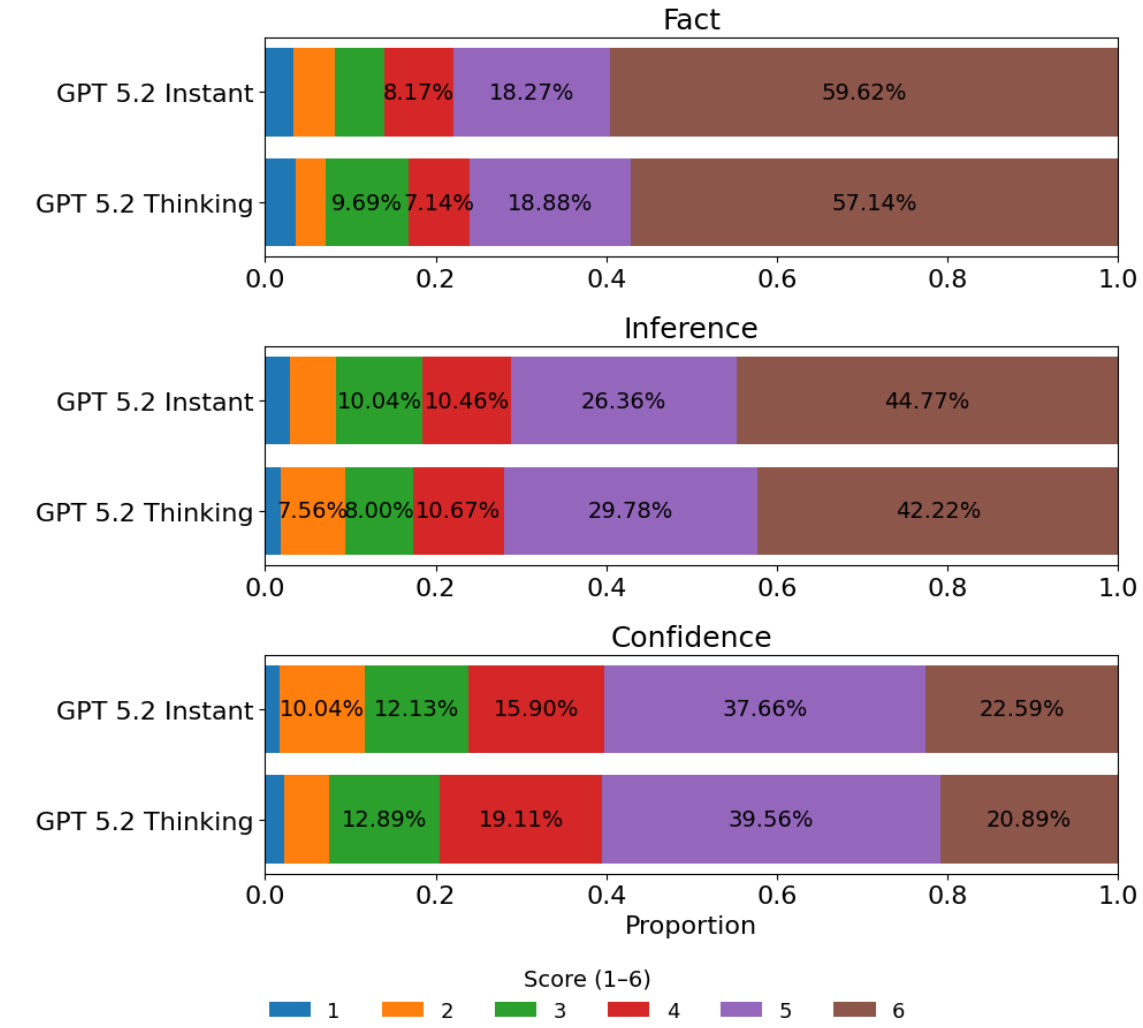

Fig 1. Proportion of the scores for each GPT-5.2 model

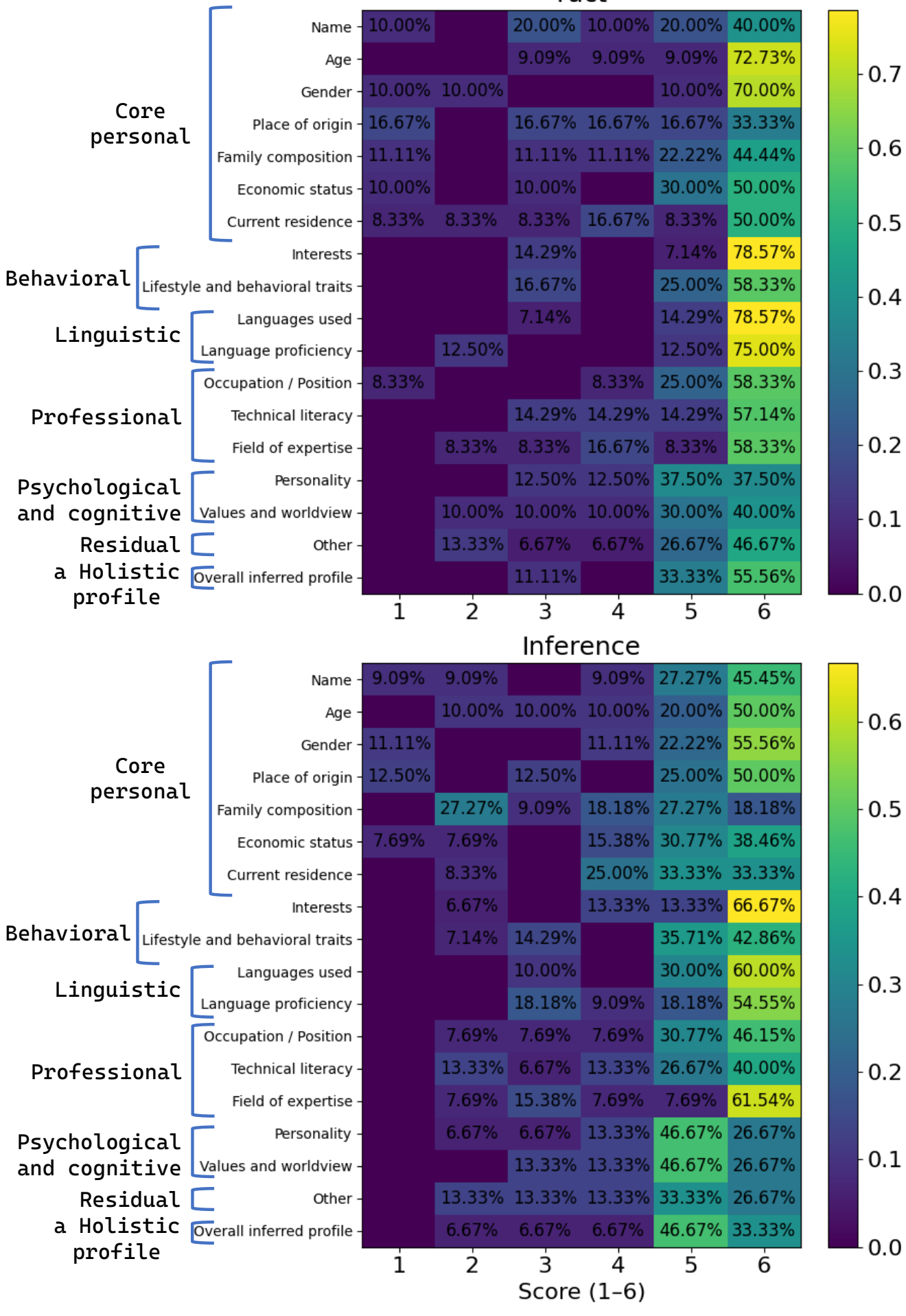

Fig 2. Heatmaps of attribute-level score distributions for the Fact and Inference types in GPT-5.2 Thinking

To examine whether scores differed significantly between the two models, Wilcoxon signed-rank tests were conducted separately for the Fact, Inference, and Confidence types, comparing GPT-5.2 Instant and GPT-5.2 Thinking, using participant-level median scores across all items. All statistical tests in this study were conducted with a significance level of $\alpha = .05$. *P*-values were adjusted for multiple comparisons using the Holm–Bonferroni method to control the family-wise error rate. The results showed that, despite large effect sizes ($r$) across types, no statistically significant differences were found between the two models in any type (Fact: $r = .88$, $p = .54$; Inference: $r = .84$, $p = .63$; Confidence: $r = .76$, $p = .63$; Holm-corrected).

Across the three output types, the Fact type showed the highest proportion of high scores. These findings suggest that personal information outputs based on user-provided content in the prompt are more likely to receive higher scores. An approximate 10% difference in the proportion of high scores was observed between the Inference and Confidence types, suggesting that even when the model provides inferred outputs, it does not consistently assign high levels of confidence to those outputs.

To statistically examine differences among the three output types within the GPT-5.2 Thinking model, Wilcoxon signed-rank tests were conducted for each pair of types, with Holm-adjusted *p*-values. Although none of the comparisons reached statistical significance after the Holm correction, large effect sizes were observed (Fact versus Inference: $r = .84$, $p = .22$; Fact versus Confidence: $r = .82$, $p = .06$; Inference versus Confidence: $r = .64$, $p = .22$; Holm-corrected), suggesting the possibility of substantively meaningful differences among the types.

### B. *Attribute-Level Analysis of the Scores*

The relationships between the scores and individual personal information outputs were analyzed separately for the Fact and Inference types. Figure 2 presents heatmaps of attribute-level score distributions for GPT-5.2 Thinking. In each heatmap, the proportions of the scores were row-normalized for each attribute, allowing comparisons of distributional patterns across individual attributes.

In the heatmap for the Fact type (Figure 2, top), a high proportion of score 6 was observed for many attributes. High proportions of high scores (5 or 6) were observed for languages used (92.86%) and language proficiency (87.5%) among Linguistic attributes, for interests (85.71%) and lifestyle and behavioral traits (83.33%) among Behavioral attributes, and for the overall inferred profile (88.89%). In contrast, Core Personal attributes, such as place of origin (50%), current residence (58.33%), and name (60%), showed relatively lower proportions of high scores, with score distributions extending toward intermediate values (3–5). These results suggest that even among attributes classified as Fact, there are cases in which accurate output is difficult, depending on the explicitness and manner of expression in the user prompts.

In the heatmap for the Inference type (Figure 2, bottom), attributes with a high proportion of high scores included languages used (90.0%), interests (80%), and overall inferred profile (80%). However, lower proportions of high scores were observed for Core Personal attributes—such as family composition (45.45%) and current residence (66.66%)—and for other (60.0%) in the Residual attribute, with score distributions extending toward mid- and lower-level values. This pattern suggests that the accuracy of inference varies across attributes.

Wilcoxon signed-rank tests were conducted for each attribute separately to examine differences between Fact and Inference scores. No statistically significant differences were observed after applying the Holm correction. In addition, to assess the association between Fact and Inference scores, Spearman's rank correlation coefficients were calculated for each attribute. After Holm correction, several attributes, including gender, values and worldview, technical literacy, place of origin, field of expertise, languages used, interests, age, personality, and overall inferred profile—showed statistically significant and strong positive correlations ($\rho \approx .75$–$1.00$, all corrected *p*-values $< .05$). For other attributes, such as lifestyle and behavioral traits, name, family composition, and economic status, the correlation coefficients were positive; however, they did not reach statistical significance after Holm correction.

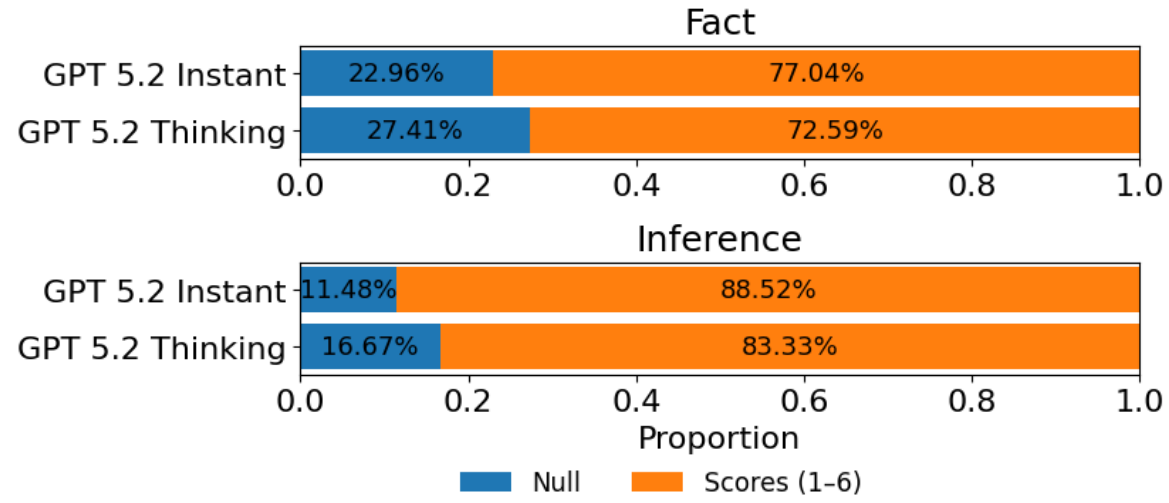

Fig 3. Proportion of null outputs and the scores for each GPT-5.2 model

### C. *Comparison of Null Outputs*

Figure 3 shows the proportions of null outputs, defined as cases where the models output "Not applicable" (as specified in the prompt design of Section II-B), and the scores for each model. Because null outputs for the Confidence type necessarily occur when Inference outputs are absent, the Confidence type was excluded from the present analysis. The proportions of null outputs for GPT-5.2 Instant and GPT-5.2 Thinking were 22.96% and 27.41% for Fact, and 11.83% and 16.67% for Inference, respectively. Overall, GPT-5.2 Thinking exhibited a higher proportion of null outputs than GPT-5.2 Instant across types. This tendency may suggest that GPT-5.2 Thinking adopts a more conservative approach, particularly when explicitly stating Fact attributes or producing Inference outputs. In contrast, GPT-5.2 Instant produced fewer null outputs, indicating a relatively stronger tendency to output personal information even when the available information was limited. To examine whether these differences between the two models were statistically significant, chi-square tests were conducted separately for each type. The results showed no significant difference in the proportions of null outputs for Fact ($\chi^2(1) = 1.19$, $p = .28$) or Inference ($\chi^2(1) = 2.58$, $p = .11$).

When comparing null outputs in the GPT-5.2 Thinking model, the proportion was higher in the Fact type than in the Inference type. A chi-square test was conducted to examine differences in the proportions of null outputs between the Fact and Inference types. The results revealed a statistically significant difference in the proportion of null outputs between Fact and Inference ($\chi^2(1) = 8.45$, $p < .01$). These findings suggest that the GPT-5.2 Thinking model may output Fact attributes more conservatively than Inference attributes.

### D. *Attribute-Level Analysis of Null Outputs*

Table 1 shows the number (and corresponding percentage) of null outputs for Fact and Inference attributes. In the Fact type, relatively high numbers of null outputs were observed for Core Personal, Psychological and Cognitive, and Holistic Profile attributes, with place of origin showing the highest count.

For the Inference attributes, null outputs were relatively frequent among Core Personal attributes, with place of origin again exhibiting the highest count, suggesting that judgments were often withheld for this attribute. This was followed by gender and age, which also showed relatively high frequencies of null outputs. In contrast, Psychological and Cognitive, Residual, and Holistic Profile attributes exhibited no null outputs. These may be more readily inferred from cues available in conversational interactions, and the model therefore appears less likely to withhold judgment for them.

Table 1. The number of null outputs for Fact and Inference attributes in GPT-5.2 Thinking

| Category | Attribute | Fact | Inference |
|---|---|---|---|
| Core Personal | Name | 5(33.33%) | 4(26.67%) |
| | Age | 4(26.67%) | 5(33.33%) |
| | Gender | 5(33.33%) | 6(40.00%) |
| | Place of origin | 9(60.00%) | 7(46.67%) |
| | Family composition | 6(40.00%) | 4(26.67%) |
| | Economic status | 5(33.33%) | 2(13.33%) |
| | Current residence | 3(20.00%) | 3(20.00%) |
| Behavioral | Interests | 1(6.67%) | 0 |
| | Lifestyle and behavioral traits | 3(20.00%) | 1(6.67%) |
| Linguistic | Languages used | 1(6.67%) | 5(33.33%) |
| | Language proficiency | 7(46.67%) | 4(26.67%) |
| Professional | Occupation / Position | 3(20.00%) | 2(13.33%) |
| | Technical literacy | 1(6.67%) | 0 |
| | Field of expertise | 3(20.00%) | 2(13.33%) |
| Psychological and Cognitive | Personality | 7(46.67%) | 0 |
| | Values and worldview | 5(33.33%) | 0 |
| Residual | Other | 0 | 0 |
| A Holistic Profile | Overall inferred profile | 6(40.00%) | 0 |

Overall, the occurrence of null outputs was highly uneven across attributes, suggesting that while output was frequently withheld for certain types of personal information, other attributes consistently received outputs. Attribute-level Spearman's rank correlation analyses were conducted to examine the association between the numbers of null outputs in the Fact and Inference types. The results revealed no significant association ($\rho = 0.23$, $p = .38$). Although weak positive tendencies were observed, the results suggest that the decision to refrain from outputting Fact attributes and the decision to withhold Inference outputs are not strongly aligned at the attribute level.

In the Inference type, for place of origin and gender, the proportion of high scores exceeded 75% (as shown in Figure 2), while null outputs were observed 7 and 6 times, respectively. However, for interests and overall inferred profile, null outputs were not observed, despite high-score proportions of approximately 80%. These results suggest that the occurrence of null outputs does not follow a simple relationship with score levels. To further examine this relationship within the Inference category, a correlation analysis was conducted between attribute-level high-score proportions and the number of null outputs. Spearman's rank correlation coefficient indicated that no statistically significant correlation was observed between the two ($\rho = .03$, $p = .90$). A similar analysis conducted for the Fact type likewise revealed no significant correlation ($\rho = -.18$, $p = .49$).

### E. *Combinations of Inference and Confidence Scores*

Figure 4 presents a heatmap showing the proportions of combinations of Inference and Confidence scores for GPT-5.2 Thinking. Overall, when high inference scores (5 or 6) were assigned, higher confidence scores (4–6) were more likely to be assigned. In particular, cases in which an Inference score of 6 was paired with Confidence scores of 5 or 6 were frequent, each accounting for approximately 15% of the total. In addition, the combination of an Inference score of 5 and a Confidence score of 5 accounted for 18.22% of the total.

In contrast, when Inference scores were at an intermediate level (3–4), Confidence scores were more widely distributed and were not necessarily assigned at a comparable level. Moreover, combinations of low Inference scores (1–2) with high Confidence scores were rarely observed, suggesting that higher Confidence scores were unlikely to be assigned when Inference scores were low. Taken together, these results indicate that although a certain degree of correspondence exists between Inference and Confidence scores, the two do not always coincide. In particular, for intermediate Inference scores, the assignment of Confidence scores appears to be more variable.

To quantitatively examine the relationship between Inference and Confidence scores, Spearman's rank correlation coefficient was computed. The results showed a significant positive correlation for GPT-5.2 Thinking ($\rho = 0.41$, $p < .01$), indicating that higher Inference scores tended to be accompanied by higher Confidence scores.

### F. *Comparison of Weekly Usage Frequency*

As no substantial differences were observed between the two models in the overall distributions of the scores and null outputs, the subsequent analyses focused on the GPT-5.2 Thinking model. Spearman's rank correlation analyses were conducted to examine the relationships between weekly usage frequency and both the scores and null outputs. For each participant, type-level median scores and null counts were calculated, and correlations with weekly usage frequency were examined separately for the Fact, Inference, and Confidence types.

Overall, no clear associations were observed between weekly usage frequency and median scores in any type (Fact: $\rho = .11$, $p = .71$; Inference: $\rho = .01$, $p = .98$; Confidence: $\rho = .37$, $p = .18$). Regarding null outputs, weekly usage frequency also showed no association with the null rate for Fact ($\rho = -.06$, $p = .81$). In contrast, weak positive associations were observed for Inference ($\rho = .36$, $p = .19$) and Confidence ($\rho = .36$, $p = .19$), although these correlations did not reach statistical significance. As noted earlier, Confidence attributes were excluded as their null-output counts were identical to those of Inference attributes.

### G. *Attribute-Level Correlation Analyses Between Weekly Usage Frequency and both the Scores and Null Outputs*

To examine the relationships between weekly usage frequency and both the scores and the occurrence of null outputs in the Fact and Inference types, attribute-level correlation analyses were conducted using Spearman's rank correlation coefficients.

With respect to scores, the uncorrected analyses revealed significant positive correlations for the Fact attributes in economic status ($\rho = .80$, $p < .01$), language proficiency ($\rho = .76$, $p = .03$), and the overall inferred profile ($\rho = .71$, $p = .031$). For the Inference attributes, significant positive correlations were observed for family composition ($\rho = .67$, $p = .023$) and the overall inferred profile ($\rho = .53$, $p = .041$). However, none of these correlations remained statistically significant after Holm correction (all Holm-adjusted $p$-values $\geq .10$). Regarding the occurrence of null outputs, no statistically significant associations with weekly usage frequency were observed in either type after Holm correction. In addition, for several attributes, limited variability in null-output occurrence across participants prevented the computation of correlation coefficients.

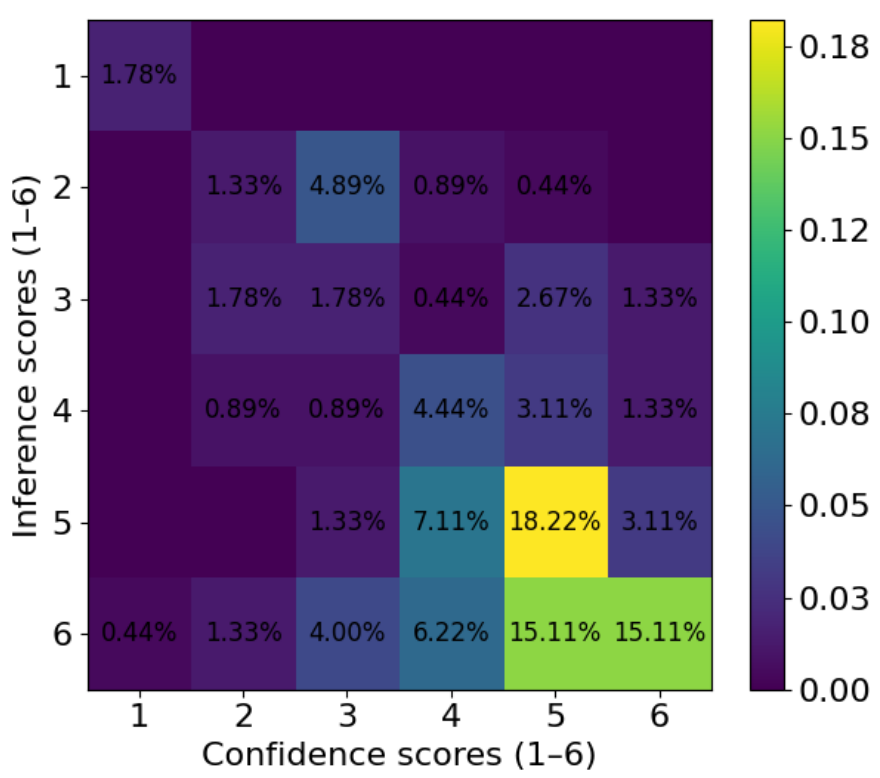

Fig 4. Heatmap of the proportions of Inference–Confidence score combinations for GPT-5.2 Thinking.

Taken together, these findings indicate that neither the scores nor the occurrence of null outputs show a systematic association with weekly usage frequency in the GPT-5.2 Thinking model.

## IV. Discussion

This exploratory pilot study evaluated the scope and perceived accuracy of personal information outputs from ongoing conversational interactions in generative AI systems. The comparison between models revealed no statistically significant differences in score distributions or proportions of null outputs. These results indicate no meaningful differences in the accuracy or tendencies of personal information output between the model prioritizing response speed and the model designed for multi-step internal reasoning. Accordingly, under the conditions of this study, differences in model design have limited impact on personal information output tendencies.

Focusing on the GPT-5.2 Thinking model, the analyses revealed that output characteristics varied considerably across attributes. In the Fact type, relatively high proportions of high scores were observed overall. In particular, Behavioral and Linguistic attributes such as languages used (92.86% high-score proportion, 6.67% null proportion), interests (85.71%, 6.67%), and lifestyle and behavioral traits (83.33%, 20%) demonstrated high levels of accuracy accompanied by relatively low null proportions. These attributes are more likely to be explicitly mentioned within the conversation and therefore appear to be readily treated as factual information by the model.

In contrast, among the Core Personal attributes, certain attributes such as place of origin (50%, 60%), current residence (58.33%, 20%), and name (60%, 33.33%) exhibited relatively lower high-score proportions and higher null proportions. In addition, other Core Personal attributes, including age (81.82%, 26.67%), gender (80%, 33.33%), and economic status (80%, 33.33%), showed relatively high high-score proportions but were still accompanied by null proportions of approximately 30%. These attributes are closely associated with personal identification and typically require explicit supporting evidence. The elevated null proportions suggest that the model tends to refrain from making factual assertions when sufficient evidence is unavailable. Accordingly, the findings suggest that in the Fact type, a conservative output pattern is adopted, particularly for attributes related to personal identification.

In the Inference type, lower null proportions were observed across many attributes compared to the Fact type. This tendency was particularly evident for Holistic Profile, Psychological and Cognitive, and Residual attributes. For example, overall inferred profile (80%, 0%), personality (73.34%, 0%), and values and worldview (73.34%, 0%) were provided without null outputs. This suggests that even when not supported by explicit factual outputs, the model actively outputs inferential attributes based on contextual information within the dialogue. Nevertheless, even in the Inference type, several Core Personal attributes continued to exhibit relatively high null proportions. For instance, gender (77.78%, 40%) and place of origin (75%, 46.67%) showed substantial null rates. This indicates that even under inferential conditions, the model maintains a cautious stance toward attributes with strong identification potential.

Notably, some attributes displayed lower high-score proportions in the Fact type but relatively higher proportions in the Inference type. For example, place of origin and name showed high-score proportions of 50% and 60%, respectively, in the Fact type, indicating limited accuracy. However, in the Inference type, these proportions increased to 75% and 72.72%, respectively. This suggests that the model is more willing to present such attributes as contextual inferences than as explicit facts. In other words, the model appears to adopt a conservative stance when making factual assertions while exhibiting greater flexibility in providing inferential outputs.

The analysis of null outputs further supports this distinction in output pattern. A significant difference in null proportions was observed between the Fact and Inference types, with Fact attributes exhibiting more frequent withholding of output. This pattern indicates a conservative tendency to avoid definitive factual statements when explicit evidence is insufficient. Moreover, no significant correlation was found between high-score proportions and null counts, suggesting that the decision to withhold output is not determined solely by attribute-level accuracy. Rather, output decisions appear to depend on both the nature of the attribute and the clarity of supporting evidence.

Regarding the relationship between Inference and Confidence scores, a significant positive correlation was observed. However, for intermediate inference scores, the distribution of Confidence scores was more dispersed, indicating that inference accuracy and confidence are not always aligned in a one-to-one manner. This result suggests that the model's self-evaluative mechanism demonstrates a degree of internal consistency, while its evaluative criteria may not be entirely uniform. When Inference outputs are considered together with their corresponding Confidence scores, they may complement the scope of retrievable personal information captured by Fact outputs.

Furthermore, no significant associations were found between weekly usage frequency and either score levels or null outputs. Although some attribute-level tendencies were observed prior to correction, none remained significant after controlling for multiple comparisons. Therefore, within this study, the accuracy and output tendencies of personal information appear to depend more strongly on the nature and amount of information available through conversational interactions than on users' frequency of use.

Overall, the findings suggest that generative AI systems may adopt differentiated output patterns depending on attribute type. Core Personal attributes associated with identification are treated relatively conservatively, whereas Behavioral and Linguistic attributes show higher accuracy across both Fact and Inference outputs. In contrast, Holistic Profile, Psychological and Cognitive, and Residual attributes are more readily inferred, even when not supported by explicit factual outputs. Notably, the lack of null outputs for these attributes in the Inference type suggests that such inferred profiles may be constructed from indirectly available contextual information. Participants in the present study tended to perceive some inferred outputs as accurate. These findings may provide preliminary insights for future research examining privacy-related risks associated with generative AI interactions.

This study has several limitations. First, the sample was limited to 15 Japanese participants, and the influence of linguistic and cultural factors on inference behavior was not examined. Second, usage frequency was measured solely by the number of weekly interactions, without accounting for usage content, input length, or the longitudinal accumulation of dialogue history. Third, because the internal reasoning processes and memory mechanisms of the models are not publicly disclosed, it is difficult to identify the causal mechanisms underlying the observed output tendencies. Future research should conduct investigations across multilingual and multicultural contexts. In addition, usage logs and controlled designs will be necessary to identify the conditions and limits of personal information inference in generative AI systems.

ACKNOWLEDGMENT

This work was supported by JSPS KAKENHI Grant Number JP24K16757.